\pdfoutput=1

\documentclass[11pt]{article}

\usepackage{url}
\makeatletter
\g@addto@macro\UrlBreaks{\do\-\do\.\do\/\do\a\do\b\do\c\do\d\do\e\do\f\do\g\do\h\do\i\do\j\do\k\do\l\do\m\do\n\do\o\do\p\do\q\do\r\do\s\do\t\do\u\do\v\do\w\do\x\do\y\do\z}
\makeatother
\usepackage{xcolor}

\usepackage[normalem]{ulem}%

\usepackage{acl}

\usepackage{times}
\usepackage{latexsym}
\usepackage[T1]{fontenc}
\DeclareTextSymbol{\textasciigrave}{T1}{0}
\usepackage[utf8]{inputenc}
\usepackage{microtype}
\usepackage{listings}
\usepackage{booktabs}
\usepackage{array}
\usepackage{tikz}
\usetikzlibrary{arrows.meta,positioning,fit}

\title{Frustratingly Easy Agentic RAG via Portable Skills: \\A Case Study in Customer Support}

\author{%
	Istok Bako \quad Bonan Min \quad Qihong (Tiffany) Shao \quad Claire Cheng \quad Yashar Mehdad \\
	Airbnb, Inc., USA \\
	\small\texttt{\{istok.bako, bonan.min, tiffany.shao, claire.cheng, yashar.mehdad\}@airbnb.com}%
}

\begin{document}
	\maketitle
	
	\begin{abstract}

		Enterprise workflow automation typically requires retrieving a complete and accurate
		set of relevant information from large operational document corpora (SOPs, policy documents)
		given a specific scenario or request. In customer support (CS), for example, human 
		specialists must determine the correct sequence of resolution steps for each case, 
		a task that demands precise grounding in policy documentation. Meeting this bar in
		production typically requires complex document chunking, indexing, and retrieval 
		pipelines, often layered with learned re-rankers, query rewriters, and bespoke 
		agentic RAG infrastructure.
		
		We report a deployment case study of a simpler alternative: a portable \emph{skill} consisting of a natural-language
		instruction, lightweight tools (a retrieval API, format-checking scripts), and a corpus 
		enriched with an LLM-generated \emph{navigation layer}, which is a structured "Wikipedia-style"
		index over the underlying documents. This agentic RAG approach is straightforward to 
		implement (1-2 engineer days), portable across capable agentic harnesses (e.g., 
		Claude Code), and performs strongly on the challenging task of CS resolution workflow
		generation, achieving 99.2\% procedural equivalence to human-approved ground truth 
		and 83.3\% evidence sufficiency. It is also roughly an order of magnitude slower per case than our deployed production assistant ($\sim$90\,s vs.\ $\sim$8--9\,s), so we claim no latency advantage. Our claim is scoped to that setting: a mature, well-curated corpus, a latency-tolerant workflow, and human review of every case. There, the skill did the work of a complex RAG stack, and maintenance became a matter of editing text and data rather than retraining models.
		
	\end{abstract}
	
	\section{Introduction}
	Customer Support (CS) aims to enhance the end-customer experience by making every interaction between customers and CS specialists smarter, faster, and more seamless.  We develop AI solutions with LLM agents that empower CS specialists to work more efficiently, effectively, and intelligently.
	
	Internal LLM agents (systems that read a work item and produce a
	structured artifact for a human operator) are increasingly common in
	enterprise settings. A representative instance is the
	\emph{customer-service guidance agent}: it ingests a
	de-identified support case, from which identifiable customer personal information has been removed,
	(contact reason, internal CS-specialist notes, and the user--specialist transcript)
	and produces step-by-step guidance for a CS specialist,
	together with verbatim supporting evidence drawn from an operational
	knowledge base.
	
	The default response is a production RAG stack: a fine-tuned or prompt-specialized generator, a multi-stage retrieval pipeline with learned re-rankers and query rewriters, and per-workflow heuristics. Such stacks deliver, but are expensive to build and, most painfully, hard to maintain as the knowledge base, contact-reason taxonomy, and user behavior drift: every task change implies a change to a model artifact or pipeline component.
	
	We describe an alternative we have found effective in practice: the
	\textbf{skill-based paradigm}, building on the idea of portable,
	model-agnostic agent \emph{skills} \cite{anthropic2025skill}. A
	\emph{skill} is a self-contained, declarative specification of an
	agentic task, comprising
	(i)~a natural-language operating manual that defines the workflow,
	constraints, and output contract; (ii)~a small set of deterministic
	scripts that the agent must call to read inputs, fetch corpus content,
	and write validated outputs; (iii)~a configuration file pinning paths
	and parameters; and (iv)~a structured corpus. The skill carries no
	model weights. It is executed by a general-purpose LLM agent harness (a
	coding-agent runtime with file access and tool use), which supplies
	the reasoning. Deploying the task means \emph{authoring a skill};
	maintaining it means \emph{editing text and corpus}, not retraining.
	
	We instantiate it for CS guidance with a single retrieval tool: hybrid (BM25 + dense) search over a structured operational wiki of \textasciitilde2{,}700 Markdown pages cross-referenced to source articles by stable identifiers, with no learned re-ranker, query rewriter, fine-tuning, or bespoke serving. The agent parses the case, runs a few queries, fetches source articles, drafts guidance, and extracts verbatim evidence, with deterministic scripts validating every step.
	
	On \textbf{132 real CS cases} from deployed traffic, this skill
	produces guidance that an LLM judge rates as equivalent to human-approved
	ground truth on \textbf{99.2\%} of cases, and supplies sufficient
	verbatim evidence to reconstruct the procedure on \textbf{83.3\%}.
	
	We are explicit about what is and is not new here. The \emph{mechanism} (an instruction-conditioned agent that calls tools and a retriever) is not novel, and commercial agent frameworks already package instructions, tools, and retrieval together. Our contributions are an LLM-generated navigation layer, a deployment lens, and an evaluation, not a new agent algorithm:
	\begin{itemize}
		\itemsep0em
		\item \textbf{An LLM-generated navigation layer} (our main contribution): generated automatically from the source articles, it combines three properties in one index: (i)~intent indexing, where each page states what its workflow solves rather than its steps~\cite{su2025llmfriendly}; (ii)~typed links to the underlying articles, a lightweight knowledge graph in the spirit of GraphRAG~\cite{edge2024graphrag} but with no graph-construction step and no multi-hop traversal at query time; and (iii)~merged entities that consolidate a procedure split across articles into a single page, eliminating the need for multi-hop retrieval. The novelty is not any single property but their automatic, jointly generated combination as a swappable retrieval substrate (Section~\ref{sec:ablations}).
		\item We frame the \textbf{skill} as a unit of \emph{deployment and maintenance} whose load-bearing property is an \emph{enforced trust boundary}: that every emitted snippet is a verbatim substring of its cited article is enforced by code rejecting non-conformant outputs, not by prompting, making the I/O contract auditable independently of the model, which is what distinguishes a ``skill'' from a system prompt with tools.
		\item We report a \textbf{deployment case study} in which the system
		is improved and maintained by editing prose and data rather than
		retraining, and quantify guidance quality on 132 real cases using a
		single hybrid retriever and no production retrieval engineering.
		\item A \textbf{two-metric evaluation} (procedural equivalence and evidence sufficiency) used as an \emph{investment localizer}: the gap between them indicates whether the next improvement belongs in the agent or the corpus, and our failure inspection (Section~\ref{sec:analysis}) attributes the residual gap to corpus content rather than retrieval or generation failures.
	\end{itemize}

	\section{Related Work}
	
	Our work builds on retrieval-augmented generation \cite{lewis2020retrieval,karpukhin2020dpr,robertson2009bm25,gao2024retrieval}, agentic retrieval and tool use \cite{yao2023react,schick2023toolformer,asai2024selfrag,wang2024agentsurvey}, the established split between answer and evidence quality in RAG evaluation \cite{es2024ragas,saadfalcon2024ares}, and LLM-as-judge evaluation \cite{zheng2023judging,stureborg2024large}. The mechanism is standard; our contribution is the \emph{deployment unit} (a script-enforced, model-agnostic skill) and the use of the procedural-equivalence/evidence-sufficiency split as an \emph{investment localizer}, not the agent loop or the metrics themselves. Appendix~\ref{app:relwork} expands these connections.
	The navigation layer is a retrieve-over-summaries index, a shape it shares with RAPTOR's recursive abstractive summarization~\cite{sarthi2024raptor}, document-summary indexes that search per-document summaries and dereference to the source text~\cite{liu2023docsummary}, and doc2query-style document expansion~\cite{nogueira2019doc2query}; Appendix~\ref{app:relwork} draws the comparison out, including with the closest work in this domain~\cite{su2025llmfriendly}. Our contribution is the deployment pattern, the maintainability evidence, and the code-enforced evidence contract, not a novel retrieval mechanism.
	
	\section{The Skill-Based Paradigm}
	\label{sec:paradigm}
	We use \emph{skill} to mean a portable, declarative package that fully
	specifies an agentic task for a general-purpose LLM agent harness, without
	embedding any task-specific model weights. A skill has four parts: an
	operating manual, a set of deterministic scripts, a configuration file
	that pins paths and parameters, and a structured corpus
	(Section~\ref{sec:system}). We describe the first two, which carry the
	design.
	\paragraph{Operating manual.}
	A natural-language document (in our case a single Markdown file,
	\texttt{SKILL.md}) that states the task, the step-by-step workflow,
	the hard constraints, and the output contract. It is read by the
	agent harness at the start of a session and functions as the agent's
	specification. Crucially it is prose, not code: it is edited by the
	operators who own the task, and version-controlled alongside the
	scripts and corpus.
	\paragraph{Deterministic scripts.}
	A small set of programs the agent must call rather than reason around: a case loader; a corpus reader returning article content by identifier; an output writer that validates against a schema and rejects non-conformant writes; and an evaluation runner. The scripts are the skill's \emph{trust boundary}: they enforce in code, not by prompting, the constraint that ``every emitted snippet is a verbatim substring of its cited source article'', exact after Unicode-punctuation and whitespace normalization, which is the only leniency the shipped validator allows (Appendix~\ref{app:scaffold}). During benchmarking the loader also strips the withheld ground-truth fields, an evaluation-integrity safeguard not present in deployment, where there is no reference.
	\paragraph{Why this is a useful unit of deployment.}
	Four properties follow: \emph{portability} (the same skill runs unchanged on any capable harness; a model upgrade is just a harness change), \emph{maintainability} (adapted by editing prose or data, not retraining), \emph{auditability} (the deterministic scripts make the I/O contract enforceable independent of the model), and \emph{separation of concerns} (retrieval, content access, generation, and validation are distinct, swappable components).
	
	\section{A Wiki-Backed Guidance Skill}
	\label{sec:system}
	
	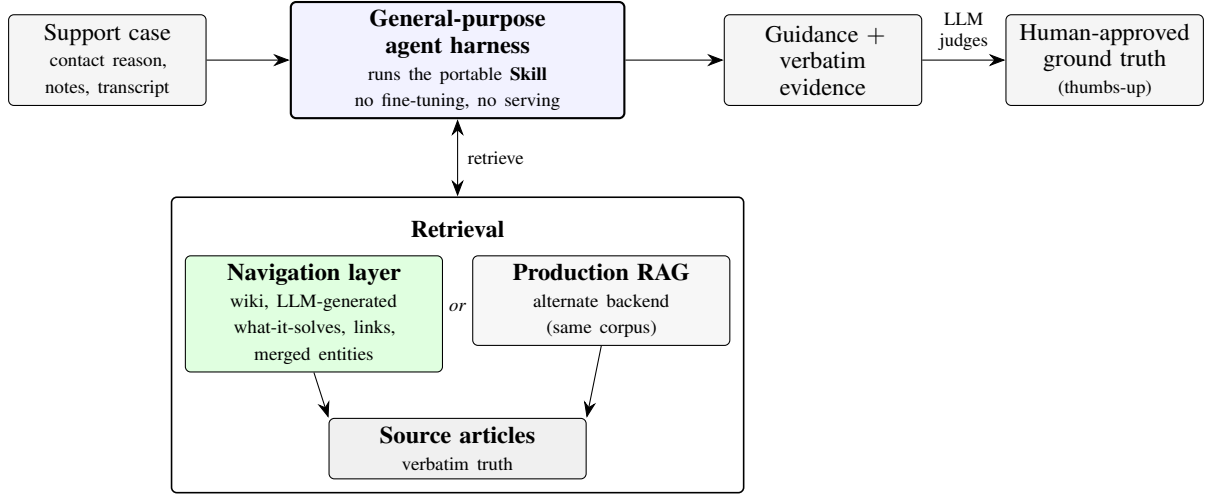
\begin{figure*}[t]
		\centering
		\begin{tikzpicture}[
			>={Stealth[length=2.2mm]}, font=\footnotesize,
			box/.style={draw, rounded corners=2pt, align=center, inner sep=3pt},
			io/.style={box, fill=black!4, minimum height=12mm, minimum width=26mm, text width=24mm},
			host/.style={box, thick, fill=blue!5, minimum height=12mm, text width=42mm},
			backend/.style={box, fill=black!3, minimum height=11mm, minimum width=34mm, text width=31mm},
			navb/.style={backend, fill=green!12},
			srcb/.style={box, fill=black!6, minimum height=8mm, minimum width=34mm, text width=31mm}
			]
			\node[host] (host) {\textbf{General-purpose agent harness}\\{\scriptsize runs the portable \textbf{Skill}}\\{\scriptsize no fine-tuning, no serving}};
			\node[io, left=11mm of host] (case) {Support case\\{\scriptsize contact reason,}\\{\scriptsize notes, transcript}};
			\node[io, right=13mm of host] (out) {Guidance $+$\\verbatim\\evidence};
			\node[io, right=11mm of out] (gt) {Human-approved\\ground truth\\{\scriptsize (thumbs-up)}};
			\node[navb, below=18mm of host, xshift=-19mm, anchor=north] (nav) {\textbf{Navigation layer}\\{\scriptsize wiki, LLM-generated}\\{\scriptsize what-it-solves, links,}\\{\scriptsize merged entities}};
			\node[backend, below=18mm of host, xshift=19mm, anchor=north] (prag) {\textbf{Production RAG}\\{\scriptsize alternate backend}\\{\scriptsize (same corpus)}};
			\path (nav.east) -- (prag.west) node[midway, font=\scriptsize\itshape] {or};
			\node[srcb, below=6mm of nav, xshift=19mm, anchor=north] (src) {\textbf{Source articles}\\{\scriptsize verbatim truth}};
			\coordinate (tc) at (host |- nav.north);
			\node[font=\footnotesize\bfseries, above=1.0mm of tc] (rettitle) {Retrieval};
			\node[draw, semithick, rounded corners=2pt, fit=(rettitle)(nav)(prag)(src), inner sep=5pt] (retbox) {};
			\draw[->] (case) -- (host);
			\draw[->] (host) -- (out);
			\draw[->] (out) -- node[above, font=\scriptsize, align=center]{LLM\\judges} (gt);
			\draw[<->] (host.south) -- node[right, font=\scriptsize]{retrieve} (host.south |- retbox.north);
			\draw[->] (nav.south) -- (src.north west);
			\draw[->] (prag.south) -- (src.north east);
		\end{tikzpicture}
		\caption{Overall architecture. A general-purpose agent harness runs the portable \emph{skill} as a fixed loop. Its \emph{Retrieval} component is swappable: the default LLM-generated wiki \emph{navigation layer} or the organization's \emph{production RAG} retriever, both surfacing the same \emph{source articles} (the two conditions ran different loops; Section~\ref{sec:results}). A deterministic validator admits only verbatim snippets; outputs are scored against human-approved ground truth by \textsc{Guidance-Match} and \textsc{Snippets-Sufficient}.}
		\label{fig:arch}
	\end{figure*}
	
	\subsection{Structured Corpus: Two Layers}
	
	The skill operates over an internal operational knowledge base
	exposed as two layers (Figure~\ref{fig:arch}).
	
	\begin{itemize}
		\itemsep0em
		\item A \textbf{wiki layer} of \textasciitilde2{,}700 Markdown pages authored as
		topic-organized summaries. Each page ends with a list of stable
		identifiers (e.g.\ \texttt{art\_NNN}) that point to the
		underlying source articles. This layer is the \emph{navigation index}:
		it is what retrieval searches over.
		\item A \textbf{source-article layer} containing the 2{,}881 underlying
		operational articles. These are the primary truth source: all
		verbatim evidence the skill emits must be quoted from here, never from
		the wiki summaries.
	\end{itemize}
	
	This separates \emph{finding} the right material from \emph{quoting} authoritative content: retrieval lands on wiki pages, then dereferences their identifiers to read source articles for evidence, and the navigation layer can be re-authored independently of the underlying articles.
	
	\paragraph{What makes the navigation layer novel.}
	The navigation pages are generated automatically from the source articles by an LLM~\cite{karpathy2026llmwiki} and combine three properties. Concretely, a single GPT-4.1 pass over the 2{,}881 source articles builds the layer via an update-or-create loop against a running page index, so merged entities and typed cross-links emerge with no separate clustering step (Appendix~\ref{app:scaffold}). First, each states \emph{what its workflow solves}, and what it does not, so queries match by intent, not procedural wording~\cite{su2025llmfriendly}. Second, each records typed links to the articles it draws on: a lightweight knowledge graph in the spirit of GraphRAG~\cite{edge2024graphrag}, but with no graph-construction or multi-hop traversal. Third, where a procedure is split across articles, a \emph{merged entity} consolidates the pieces into one page, so a need that would otherwise require multi-hop retrieval is met by a single page, moving the difficulty from the retriever into one-time, LLM-authored content.

	\paragraph{Navigation-layer quality.}
	Because the layer is generated rather than curated, we audit it directly. An independent, stronger model judged every one of the 3{,}168 navigation-to-source links (the complete link set, not a sample) against the articles they cite: approximately 96\% were relevant and source-supported, while roughly 4\% were dangling, irrelevant, or insufficiently supported; the flagged links were corrected in a follow-up QA pass. A structural check confirms that all 2{,}881 source articles are referenced by at least one navigation page, so no source content is orphaned at the identifier level (Appendix~\ref{app:scaffold}). The layer is navigational, not authoritative: a weak link costs recall, never evidence integrity, because every emitted span is dereferenced from a cited source article and re-checked at save time (Section~\ref{sec:analysis}).
	
	\subsection{Retrieval Tool}
	
	The only retrieval component is a single hybrid search tool over the wiki layer, combining BM25 \cite{robertson2009bm25} with dense vector similarity \cite{karpukhin2020dpr}; it exposes keyword, hybrid, and direct-page fetch, with no learned re-ranker, query rewriter, or per-workflow heuristic. It is swappable: any retriever mapping a query to ranked wiki pages, and thence to source-article identifiers, satisfies the skill's contract.
	
	\subsection{Agent Loop}
	
	For each case the agent harness executes a fixed loop defined by the operating manual: it parses the case; issues three to five hybrid queries and aggregates the cited source-article identifiers; fetches the article bodies; drafts guidance; extracts up to $k$ verbatim spans ($k=15$); and validates and saves, with the writer re-checking every snippet against its cited article (Appendix~\ref{app:scaffold}). Per case this is a median of roughly five queries and six fetches; the agent never sees prior outputs or the ground truth.
	
	\section{Evaluation}
	
	\subsection{Cases and Inputs}
	
	We evaluate on a fixed benchmark of \textbf{132 real CS cases} sampled from deployed traffic. For each case, personally identifiable information is removed and the agent is then given the contact reason, the internal CS-specialist notes, and the user--specialist transcript. Each case carries a human-approved \textbf{ground-truth guidance document}: suggestions, steps, tips, and relevant articles a CS specialist reviewed and approved (thumbs-up) as correct. We use it as the validation reference. It is withheld from the agent at inference time and used only by the evaluation scripts.
	
	\paragraph{Ground-truth provenance.} 	The ground-truth guidance was approved (thumbs-up) by experienced CS specialists as part of the organization's existing quality-assurance process, before the agent existed; we reuse it rather than commissioning new annotations. We did not re-audit it; cases whose ground truth is itself debatable score as failures regardless of agent quality, biasing our numbers \emph{downward}.
	
	\paragraph{Why parametric knowledge cannot substitute for retrieval.}
	\label{sec:eval-retrieval}
	The corpus is internal operational content: tool navigation, routing rules, and version-specific policy, none of it likely to be well represented in a general-purpose model's pretraining. More decisively, \textsc{Snippets-Sufficient} cannot be satisfied without retrieval \emph{by construction}: a passing case requires verbatim substrings of internal articles, which the writer script enforces, so the agent must fetch them. The 83.3\% is therefore itself evidence that retrieval is load-bearing, not a number a no-retrieval baseline could approach; Section~\ref{sec:ablations} isolates the retrieval substrate and the layer's content organization separately.
	
	\subsection{Metrics}
	
	Each case is scored on two LLM judge metrics.
	
	\paragraph{\textsc{Guidance-Match} (procedural equivalence).}
	The judge sees the agent's guidance and the ground-truth guidance and
	decides: \emph{could a CS specialist use the candidate guidance to
		resolve this case equivalently to the ground truth?} The check
	tolerates paraphrase and additional adjacent procedures; it scores
	procedural outcome and semantic meaning, not surface text.
	
	\paragraph{\textsc{Snippets-Sufficient} (evidence reconstruction).}
	The judge sees the case, the ground-truth guidance, and the agent's
	verbatim snippets with their source-article identifiers, and decides:
	\emph{could a CS specialist reconstruct guidance equivalent to the
		ground truth using only these snippets?} This is strictly harder: it
	requires the procedural text the ground truth cites to exist verbatim
	in one of the cited source articles.
	
	Both metrics are evaluated by majority vote over \textbf{three
		independent judge invocations} \cite{zheng2023judging,es2024ragas}.
	We observed non-trivial single-invocation disagreement, on the order
	of 15\%, and higher on the harder \textsc{Snippets-Sufficient} metric
	than on \textsc{Guidance-Match} \cite{stureborg2024large}.
	Majority-voting over three runs is what we report; we estimate
	residual instability at roughly $\pm$2--3 percentage points, well
	below the 16-point gap between our two metrics. 

	\paragraph{Models and decoding.}
	The guidance generator is Claude Opus~4.7, run in Claude Code with adaptive thinking (\texttt{xhigh}), provider-default temperature and top-$p$, and a 32{,}000-token output cap. The navigation-layer builder and the primary judge are both OpenAI GPT-4.1 at temperature~0, with top-$p$ omitted and a 16{,}384-token completion cap; every judge decision is a majority vote over three independent calls. The dated snapshot behind the \texttt{gpt-4.1} alias was not pinned at run time and cannot be recovered retrospectively. Generator and judge are therefore from different vendors and model families, so the headline numbers are not a same-family self-evaluation. The judge sees the case, the ground-truth guidance, and the candidate guidance or its snippets, but never the navigation pages, so generated page text cannot leak into a judge decision. As supporting evidence for the judge design, an orthogonal internal effort separately calibrated a comparable GPT-4.1 judge across groundedness, completeness, policy alignment, and absence of redundancy.
	
	\section{Results}
	\label{sec:results}

	\begin{table}[t]
		\centering
		\small
		\begin{tabular}{lcc}
			\toprule
			\textbf{Metric} & \textbf{Pass rate} & \textbf{95\% CI} \\
			\midrule
			\textsc{Guidance-Match}      & 99.2\% & [95.8, 99.9] \\
			\textsc{Snippets-Sufficient} & 83.3\% & [76.1, 88.7] \\
			\bottomrule
		\end{tabular}
		\caption{Pass rates on the 132-case benchmark (131/132 and 110/132 cases), majority-voted over
			three independent judge invocations; 95\% Wilson confidence intervals~\cite{wilson1927}.}
		\label{tab:results}
	\end{table}
	
	Table~\ref{tab:results} reports the headline numbers: with a minimal retrieval layer and no model training, the skill reaches 99.2\% procedural equivalence and 83.3\% evidence sufficiency. \textsc{Snippets-Sufficient} is necessarily lower: over the \emph{same} output it asks the strictly harder question, requiring the procedure to be reconstructable from the agent's verbatim spans, which holds only when the needed text exists quotably in the cited articles. The 16-point gap is thus cases with equivalent guidance that could not be fully grounded in quotable evidence, a shortfall we trace (Section~\ref{sec:analysis}) to missing corpus content, not retrieval or generation. Because \textsc{Guidance-Match} is reference-based and evaluated without a no-retrieval baseline, part of the 99.2\% may reflect generically-reasonable rather than case-specific guidance; we therefore treat \textsc{Snippets-Sufficient}, which retrieval provably gates (Section~\ref{sec:eval-retrieval}), as the more informative metric.
	
	\paragraph{Head-to-head: two retrieval workflows.}
	Because the retriever is meant to be swappable
	(Section~\ref{sec:paradigm}), we swapped the hybrid wiki search for the
	organization's production RAG retrieval service over the same corpus, holding the corpus, the 132 cases, and the evaluation protocol fixed, but \emph{not} the query and fetch budgets or the agent's behavior: the wiki condition ran the full agent loop, whereas the production-RAG condition used a scripted query--retrieve--compose pipeline, and ran the full
	132-case benchmark (Table~\ref{tab:headtohead}). Both backends reach near-ceiling procedural equivalence (99.2\% vs.\ 96.2\%), but the lightweight navigation layer is markedly stronger on the harder, retrieval-gated evidence metric (83.3\% vs.\ 59.8\%). The \textsc{Snippets-Sufficient} gap shows as non-overlapping 95\% Wilson intervals (wiki $[76.1, 88.7]$ vs.\ production RAG $[51.3, 67.8]$); the \textsc{Guidance-Match} intervals overlap ($[95.8, 99.9]$ vs.\ $[91.4, 98.4]$), consistent with the near-tie.

	We read this as an end-to-end comparison of two operating points, not as evidence that the retrieval representation alone explains the gap. Both backends search the \emph{same} corpus, so the needed text is reachable by either; what differs is the workflow that surfaces it. The navigation layer states what each workflow solves and links to the exact underlying articles, and, because the layer is meant to be traversed, the agent can issue a targeted follow-up search once it identifies missing evidence. That re-retrieval is a property of the layer's design rather than a larger fixed budget: the production workflow completed retrieval before composition and so could not re-retrieve. Adding that behavior to the scripted baseline would require extending its orchestration; the skill expresses it in prose in the operating manual, within the agent-and-tool loop it already runs. Part of the evidence gap plausibly reflects this difference, but we do not claim the representation alone accounts for it, and the available artifacts cannot establish whether the gap survives at matched query and fetch effort (see \hyperref[sec:limitations]{Limitations}).
	
	\begin{table}[t]
		\centering
		\small
		\begin{tabular}{lcccc}
			\toprule
			\textbf{Retrieval backend} & \textbf{Qry.} & \textbf{Src.} & \textbf{Guid.} & \textbf{Evid.} \\
			\midrule
			Wiki backend (ours)  & 5 & 6 & 99.2\% & 83.3\% \\
			Production RAG  & 5 & 10 & 96.2\% & 59.8\% \\
			\bottomrule
		\end{tabular}
		\caption{Full 132-case head-to-head with the retrieval backend
			swapped (same corpus, cases, and judge protocol; query and fetch budgets and agent behavior \emph{not} held constant): \textsc{Guidance-Match} and
			\textsc{Snippets-Sufficient} pass rates, with median retrieval queries (Qry.) and median distinct sources (Src.) per case. The two source measures are \emph{not} directly comparable: the wiki figure counts source articles the agent fetched, whereas the production figure counts distinct sources represented in its emitted evidence, not service fetches. The production backend was the organization's deployed search service, called at its unified-search endpoint in the configuration used by the production assistant but driven by a scripted query--retrieve--compose harness rather than by the assistant itself; it retrieved before composition and so could not re-retrieve once composition had begun. Neither the service nor its index was tuned to these cases, and its internal retrieval depth, lexical/dense weighting, query rewriting, reranking, and index freshness are not exposed and cannot be characterized from the available artifacts.}
		\label{tab:headtohead}
	\end{table}

	\subsection{Ablations}
	\label{sec:ablations}

	Because the navigation layer is the paper's main contribution, we isolate it in two ways on the same 132 cases, holding the skill, harness, model, judge protocol, and $k=15$ fixed and varying one factor per arm (Table~\ref{tab:ablations}). We score on \textsc{Snippets-Sufficient}, the retrieval-gated metric; \textsc{Guidance-Match} is at ceiling and uninformative here.

	Varying only the search method moves little: BM25 alone reaches 103/132 and dense alone 104/132, against 110/132 for hybrid. Varying the layer's \emph{content organization} moves more. Removing merged-entity consolidation is the largest single drop (101/132), removing the intent summaries costs 104/132, weakening the typed links costs 107/132, and retrieving over the raw source articles with no navigation layer at all gives 103/132. Consolidation removed scores slightly \emph{below} no layer at all, which we read as fragmentation: forcing one page per source article leaves the layer interposed between query and article while removing the property that made it worth traversing.

	We report these as \emph{directional}. Every ablated arm's Wilson interval overlaps the full system's, and exact paired McNemar tests~\cite{mcnemar1947} over the per-case outcomes separate none of them from the full system at $n=132$: the smallest $p$ is 0.078, for removing merged entities, and the search-method arms are firmly null ($p=0.21$ and $p=0.26$). Per-arm discordant counts and $p$-values are in Appendix~\ref{app:ablations}. The pattern we take from this is narrow but useful for a practitioner: on this corpus the value lies in how the retrieval surface is organized rather than in the search algorithm applied to it, and no single property of the layer accounts for that on its own.

	\begin{table}[t]
		\centering\small
		\setlength{\tabcolsep}{3pt}
		\begin{tabular}{lcccr}
			\toprule
			\textbf{Condition} & \textbf{SS} & \textbf{\%} & \textbf{95\% CI} & \textbf{$\Delta$} \\
			\midrule
			\multicolumn{5}{l}{\emph{Retrieval substrate (same navigation layer)}} \\
			Hybrid (default)   & 110/132 & 83.3 & [76.1, 88.7] & -- \\
			BM25 only          & 103/132 & 78.0 & [70.2, 84.2] & $-5.3$ \\
			Dense only         & 104/132 & 78.8 & [71.1, 84.9] & $-4.5$ \\
			\midrule
			\multicolumn{5}{l}{\emph{Content organization (same hybrid retriever)}} \\
			Full layer & 110/132 & 83.3 & [76.1, 88.7] & -- \\
			$-$ merged entities        & 101/132 & 76.5 & [68.6, 82.9] & $-6.8$ \\
			$-$ intent summaries       & 104/132 & 78.8 & [71.1, 84.9] & $-4.5$ \\
			$-$ typed links & 107/132 & 81.1 & [73.5, 86.8] & $-2.3$ \\
			No layer (raw articles) & 103/132 & 78.0 & [70.2, 84.2] & $-5.3$ \\
			\bottomrule
		\end{tabular}
		\caption{Ablations on the same 132 cases, scored on
			\textsc{Snippets-Sufficient} (SS) with the skill, harness, model, judge
			protocol, and $k=15$ held fixed. Pass counts are given as $k/132$ with
			the corresponding pass rate; both the rate and the interval are
			percentages. The typed-link arm weakens that field rather than
			removing it (Appendix~\ref{app:ablations}). $\Delta$ is the paired difference from
			the full system in percentage points; intervals are 95\% Wilson. All
			ablated intervals overlap the full system's, so we read these as
			directional: exact paired McNemar tests separate no arm from the
			full system (Table~\ref{tab:mcnemar}).}
		\label{tab:ablations}
	\end{table}

	\section{Analysis}
	\label{sec:analysis}
	
	\textbf{The verbatim validator is the load-bearing guarantee.} Every emitted snippet is checked at save time to be a verbatim substring of its cited article (modulo Unicode-punctuation normalization), and non-conformant writes are rejected, so the evidence contract is a property of the deterministic trust boundary (Section~\ref{sec:paradigm}), not of model behavior. In the reported run, 99.7\% of emitted snippets were byte-identical to their cited article before normalization and 100\% satisfied the enforced, normalization-tolerant contract; the residual 0.3\% differed only by a curly-versus-straight apostrophe. The denominator is the 1{,}769 snippets emitted across the 132 cases of that run; validator rejection and retry counts were not logged.
	
	\paragraph{The gap is consistent with missing content, not retrieval
		or generation.}
	We inspected the 22 cases that fail \textsc{Snippets-Sufficient}, a qualitative inspection by the authors, not an adjudicated annotation, so we report it as a supporting observation without per-category counts (Appendix~\ref{app:failures}). The dominant pattern: the agent fetched relevant articles and its guidance is rated equivalent to ground truth, yet a specific procedural element the ground truth depends on is not present verbatim in those articles: a \emph{corpus-content gap}, not a retrieval failure. Typically that element is an action (an internal cross-check, fallback step, tool-navigation path, or escalation target) known to experienced CS specialists but unwritten in any source article. As a concrete (synthetic) instance, a case may hinge on a fallback step, \emph{``if the file-upload step fails, fall back to the email intake channel''}, that experienced specialists apply routinely but that is not written in the fetched articles (the guidance is equivalent, yet no verbatim snippet supports that step).
	
	Closing gaps of this kind requires \emph{authoring or extending source articles}, not improving retrieval, generation, or the skill; Appendix~\ref{app:failures} gives synthetic illustrations of their shape.
	
	\paragraph{The single guidance failure.}
	Exactly one case fails \textsc{Guidance-Match} (one of three judges accepted): a \emph{wrong-state assumption}, where the case admits two procedural phases and the agent produced guidance for the wrong one because the input under-specified which applied. At $n=1$ this is illustrative of a failure \emph{mode}, not a meaningful rate; it suggests explicit case-state enumeration in the operating manual as a mitigation.

	\section{Discussion}
	\label{sec:discussion}

	\paragraph{Scope and generality.}
	Maintainability is a \emph{design property} of this factoring rather than a measured result: because no task competence is baked into model weights, adapting the task is an edit to prose or data rather than a retraining cycle. We do not claim production pipelines are obsolete: the skill says nothing about latency, cost-per-case, or robustness under load, it operates over a mature, well-curated corpus, and learned re-rankers may help more on noisier or younger corpora, where content gaps would shift the bottleneck to authoring. The evaluation design generalizes: any task with a verifiable ground-truth procedure and a citable corpus admits the split between procedural equivalence and evidence reconstruction, and the gap between them indicates whether the next investment belongs in the agent or the corpus. We reproduce the skill in Appendices~\ref{app:scaffold} and~\ref{app:package}: operating manual, script interfaces, output schema, the validator that enforces the evidence contract, and the structure of the judge prompts, with no cases and no article text.

	\paragraph{Maintaining the navigation layer.}
	Because the layer is generated, it drifts as the source corpus changes. Maintenance is three steps, none of which touches a model: structural checks that surface broken or dangling references, an independent semantic audit that flags weakly supported links (Section~\ref{sec:system}), and removal or regeneration of the flagged pages after a corpus update. This is the sense in which the burden moves to text and data curation: it is a real burden, not an absent one.
	
	Section~\ref{sec:ablations} reports these ablations: the search method is within noise, while the layer's content organization is not.
	
	\section{Conclusion}
	
	We presented the skill-based paradigm for deploying internal LLM agents: a declarative, model-agnostic package of operating manual, deterministic scripts, configuration, and corpus, run by a general-purpose harness with no task-specific training. A wiki-backed instantiation for CS guidance, with a single hybrid retriever and no production retrieval engineering, reaches 99.2\% guidance equivalence and 83.3\% evidence sufficiency on 132 real cases. Maintaining it by editing text and data rather than retraining is our practical deployment lesson for mature operational corpora. 
	
	\phantomsection\label{sec:limitations}
	\section*{Limitations}
	
	\textbf{Scope of the study.} Our study is a focused single-deployment case study, and we state results within that scope. The benchmark is 132 real cases over one mature, well-curated corpus; we expect the recipe to transfer best to similarly mature corpora, while younger or sparser ones may benefit more from additional retrieval engineering. All evidence here comes from one internal benchmark on one corpus; we make no cross-domain superiority claim, and a faithful port to a public procedural-support benchmark remains future work.%
	
	\textbf{Evaluation.} Quality is scored by LLM judges with majority voting over three runs (estimated residual uncertainty $\pm$2--3 points); human re-scoring and a second judge family are natural complements we leave to future work. The head-to-head compares two end-to-end operating points rather than two retrievers at matched effort: query and fetch budgets were not held constant, and the artifacts we retain cannot establish whether the evidence gap survives at matched navigation effort.
	
	\textbf{Scope of the contribution.} Our claims concern the skill as a deployment unit rather than retriever internals: we show that a single hybrid retriever suffices and that retrieval is load-bearing (Section~\ref{sec:analysis}). Section~\ref{sec:ablations} isolates the retrieval substrate and the layer's content organization, but at $n=132$ those differences are directional rather than significant, and each property-level claim is limited to the specific operationalization we ablated; and because we optimize for maintainability rather than latency (Table~\ref{tab:latency}), throughput-sensitive deployments would add standard batching, caching, or parallel fetch.

	\begin{table}[t]
		\centering
		\small
		\setlength{\tabcolsep}{3pt}
		\begin{tabular}{lccc}
			\toprule
			\textbf{Backend} & \textbf{Search} & \textbf{Generation} & \textbf{Total} \\
			\midrule
			Wiki backend (ours)  & \textasciitilde1\,s    & \textasciitilde89\,s   & \textasciitilde90\,s \\
			Production RAG        & \textasciitilde5\,s    & \textasciitilde69\,s   & \textasciitilde74\,s \\
			Production assistant  & \textasciitilde3--4\,s & \textasciitilde4--5\,s & \textasciitilde8--9\,s \\
			\bottomrule
		\end{tabular}
		\caption{Per-case latency (medians): retrieval/search vs.\ the multi-step generation/agentic loop. Search is negligible for the skill (served warm); the wiki backend runs more navigation steps during generation than the production-RAG backend. The deployed production assistant's \textasciitilde8--9\,s splits between retrieval (\textasciitilde3--4\,s) and generation (\textasciitilde4--5\,s).}
		\label{tab:latency}
	\end{table}
	
	\section*{Ethical Considerations}
	
	The 132 cases involve real customer-support interactions and are
	subject to the organization's data-handling policies. No real
		case content, customer data, or personally identifying information is
		reproduced anywhere in this paper. The example strings in
	Section~\ref{sec:analysis} are \emph{synthetic}: they were written by
	the authors purely to illustrate the \emph{shape} of a failure
	category, and do not reproduce any real procedure, article, customer
	utterance, or internal entity name. Quantitative results are reported
	only as aggregate pass rates and counts. The skill runs against
	internal infrastructure and does not transmit customer data to
	external systems. Its outputs are intended to assist human CS specialists, not to act autonomously on customer accounts; deploying such a
	skill in a fully autonomous setting would require safeguards beyond
	those discussed here.

	\appendix
	
	\section{Extended Related Work}
	\label{app:relwork}
	
	\paragraph{Retrieval-augmented generation.}
	RAG \cite{lewis2020retrieval} with dense retrieval \cite{karpukhin2020dpr} and BM25 \cite{robertson2009bm25} is standard practice; production stacks add learned re-rankers and query expansion \cite{gao2024retrieval}. We instead use a single hybrid retriever as one swappable component, omitting that machinery to isolate what the corpus and a general-purpose harness deliver on their own.
	
	\paragraph{Retrieve-over-summaries indexes.}
	Retrieving over generated summaries and dereferencing to source text is an established pattern: RAPTOR clusters and recursively summarizes chunks into a tree and retrieves at a chosen level of abstraction~\cite{sarthi2024raptor}; document-summary indexes retrieve documents by summary relevance and then return their underlying nodes~\cite{liu2023docsummary}; doc2query expands documents with predicted queries so that lexical retrieval matches how users ask~\cite{nogueira2019doc2query}. Our navigation layer differs in what the summary layer is \emph{for}. It is authored to be navigated rather than to be a compressed substitute for the source: each page states what its workflow solves and what it does not, typed links name the articles it draws on, and a procedure split across articles is consolidated onto one page. Crucially, no summary is ever quoted: all verbatim evidence is dereferenced from the cited source articles and checked in code (Section~\ref{sec:paradigm}), so the summary layer carries navigation risk but never evidence risk.

	\paragraph{Knowledge representation for customer support.}
	Closest to our setting, \citet{su2025llmfriendly} restructure a customer-support knowledge base into an LLM-friendly representation and motivate indexing by what a workflow solves, which is the property our intent summaries adopt. We differ in two ways that matter for deployment rather than for retrieval quality: the source articles remain the sole evidence authority, so the generated layer can be regenerated at will without touching what may be quoted; and the resulting contract is enforced by a validator rather than requested in a prompt, which is what makes the boundary auditable independently of the model.

	\paragraph{Agentic retrieval and tool use.}
	ReAct \cite{yao2023react}, tool-use frameworks \cite{schick2023toolformer}, and self-reflective retrieval \cite{asai2024selfrag} interleave reasoning, retrieval, and tool calls \cite{wang2024agentsurvey}. Our loop is a deliberately plain instance; the contribution is the \emph{deployment} unit, not the loop: a declarative, model-agnostic skill whose hard constraints are script-enforced (Section~\ref{sec:paradigm}).
	
	\paragraph{RAG evaluation.}
	Separating answer quality from evidence quality is established \cite{es2024ragas,saadfalcon2024ares}. Our two-metric design instantiates this split; the contribution is its use as an \emph{investment localizer}, reading the gap to decide whether the next effort belongs in the agent or the corpus, with a failure analysis attributing the gap.
	
	\paragraph{LLM-as-judge evaluation.}
	We evaluate with LLM judges \cite{zheng2023judging}, mitigating their variance \cite{stureborg2024large} by majority-voting over three independent invocations.
	
	\section{Failure Inspection}
	\label{app:failures}
	
	The inspection underlying Section~\ref{sec:analysis} was qualitative.
	We examined the 22 cases failing \textsc{Snippets-Sufficient} and,
	for each, checked whether the articles the ground truth references had
	been fetched by the agent and whether the procedural element the
	ground truth depends on appeared verbatim in those articles. The
	inspection was performed by the authors, used only article identifiers
	and the presence or absence of verbatim spans, and examined no
	customer data.

	We deliberately do not report an adjudicated
	per-category count: the inspection was not a multi-annotator
	annotation with an agreement statistic, and we present it only as a
	qualitative indication of where the residual gap lies.

	A formal,
	independently labeled failure taxonomy is future work. 	All illustrative examples in this paper are synthetic (see Ethical Considerations); no real corpus or case strings are reproduced.
	
	Synthetic illustrations of the shape of these gaps (no real procedure, article, or data): ``re-confirm the adjustment amount against the counterparty's written agreement before issuing''; ``if the file-upload step fails, fall back to the email intake channel''; ``open the internal billing console, locate the transaction, copy its reference ID''; ``route to the specialist queue when a safety flag is present.''
	
	\section{Ablation Details}
	\label{app:ablations}

	All arms in Table~\ref{tab:ablations} run over the same 132 cases with the skill, harness, generator, judge prompts, judge protocol (three-call majority vote), and $k=15$ fixed; exactly one factor changes per arm. The retrieval-substrate arms reuse the navigation layer unchanged and vary only the search method. The content-organization arms rebuild the layer with one property removed: merged entities disabled by forcing one page per source article, intent summaries replaced by raw procedural text in the indexed field, and the typed links weakened rather than dropped: every page-to-source edge is preserved, but source identifiers are moved out of their inline and section-local positions into a single deduplicated page-level pool, so that arm tests local claim-to-source association rather than the presence of links. Because all arms share the same cases, the reported $\Delta$ is the mean paired difference from the full system.

	\paragraph{Paired tests.}
	Because each arm is scored on the same cases as the full system, the informative comparison is the discordance: how many cases pass under the full system but fail under the arm ($b$), and how many do the reverse ($c$). Table~\ref{tab:mcnemar} reports both, with two-sided exact McNemar $p$-values~\cite{mcnemar1947}. We use the exact binomial form on the discordant pairs rather than the asymptotic $\chi^2$ approximation, which is the appropriate choice at discordance counts of 19--23. No arm separates from the full system at the 0.05 level. Removing merged entities is the closest ($b=15$, $c=6$, $p=0.078$), and the two search-method arms are furthest from significance despite discordance of 20--23 cases, which is what a genuine null looks like: the arms disagree on individual cases in both directions in roughly equal measure. We therefore report the deltas as directional and make no property-level causal claim.

	\begin{table}[t]
		\centering\small
		\setlength{\tabcolsep}{4pt}
		\begin{tabular}{lccc}
			\toprule
			\textbf{Arm} & \textbf{$b$} & \textbf{$c$} & \textbf{Exact $p$} \\
			\midrule
			BM25 only            & 15 & 8 & 0.210 \\
			Dense only           & 13 & 7 & 0.263 \\
			$-$ intent summaries & 14 & 8 & 0.286 \\
			$-$ typed links      & 11 & 8 & 0.648 \\
			$-$ merged entities  & 15 & 6 & 0.078 \\
			No layer             & 14 & 7 & 0.189 \\
			\bottomrule
		\end{tabular}
		\caption{Paired discordance against the full system on
			\textsc{Snippets-Sufficient}: $b$ is cases the full system passes and the
			arm fails, $c$ the reverse. $p$ is the two-sided exact McNemar test. Each
			arm's net delta is $b-c$, matching Table~\ref{tab:ablations}.}
		\label{tab:mcnemar}
	\end{table}

	\paragraph{Evidence-coverage correlate.}
	As an uncontrolled observation, cases issuing a refinement query (for a sub-topic uncovered by the first round) or an entity-triggered query tended to have higher snippet coverage than single-query cases. We report this as a hypothesis, not a measured effect; notably, such behaviors are prose edits to the operating manual, needing no model or retriever change.
	
	\section{Skill Scaffold}
	\label{app:scaffold}
	For reproducibility we reproduce the skill in this appendix and the next, with no cases, no article text, and no organization-specific names: the operating manual (\texttt{SKILL.md}); the four deterministic script interfaces below, with the output schema (\texttt{scripts/INTERFACES.md}); a runnable, data-free validator that enforces the schema and the verbatim-evidence constraint (\texttt{scripts/validate\_output.py}); and a structural description of the evaluation-judge prompts, covering guidance correctness and evidence sufficiency. The validator takes an output file and a mapping from article identifier to article body, so it runs on any corpus.
	
	\paragraph{Navigation-layer build.}
	The navigation layer is generated by an incremental, index-aware ingestion loop driven by a single general-purpose LLM (GPT-4.1), one pass per source article. For each of the 2{,}881 articles the builder prompts the model with (i)~a fixed page schema, (ii)~the current \texttt{index.md} (the running list of pages already created, which serves as the wiki's memory), and (iii)~the article text (id, title, source, content). The model identifies the article's topics, entities, and procedures; updates an existing page when topics overlap or creates a new one otherwise; emits the full Markdown page per the schema; and updates \texttt{index.md} and a \texttt{log.md}. Merged entities and the typed cross-links are emergent from this update-or-create loop rather than a separate clustering pass, and the loop retries failed articles until all are covered: fully automatic, with no human in the loop.

	\paragraph{Navigation-layer audit.}
	To check the generated layer we extracted every navigation-to-source link (3{,}168, the complete set rather than a sample) and asked an independent, stronger model than the builder to judge each link against the article it cites, scoring whether the cited article supports the procedure the linking page attributes to it. Approximately 96\% passed; the roughly 4\% that did not were dangling, irrelevant, or insufficiently supported, and are removed or regenerated as part of maintenance (Section~\ref{sec:discussion}). Separately, a structural pass confirmed that every one of the 2{,}881 source articles is referenced by at least one navigation page.
	
	\paragraph{Operating manual (\texttt{SKILL.md}).}
	A prose specification read at session start, stating the task; the six-step workflow (parse, retrieve, fetch, draft, extract evidence, validate-and-save; Section~\ref{sec:system}); the hard constraints (every emitted snippet must be a verbatim substring of its cited source article); and the output contract (a guidance document plus up to $k$ source-tagged verbatim snippets).
	
	\paragraph{Script interfaces.}
	\begin{itemize}
		\itemsep0em
		\item \texttt{load\_case(id)} returns the case fields (contact reason, CS-specialist notes, transcript) and strips any reference/ground-truth fields.
		\item \texttt{retrieve(query)} returns ranked navigation pages and the source-article identifiers they cite.
		\item \texttt{get\_article(id)} returns the raw source-article body.
		\item \texttt{save\_output(guidance, snippets)} validates and writes the output, or rejects it (below).
	\end{itemize}
	
	\paragraph{Verbatim validator (enforced at save).}
	For each emitted snippet $s$ citing article $a$: normalize Unicode punctuation in $s$ and in \texttt{get\_article}($a$), and accept only if $s$ is a substring of $a$; any non-conformant snippet rejects the entire write. This pins emitted evidence to verbatim source text independent of the model.
	
	\paragraph{Judge prompts.}
	Two binary LLM-judge prompts, each majority-voted over three runs; Appendix~\ref{app:package} describes their structure. \textsc{Guidance-Match}: given the candidate guidance and the ground truth, decide whether a CS specialist could use the candidate to resolve the case equivalently to the ground truth (tolerating paraphrase and additional adjacent procedures). \textsc{Snippets-Sufficient}: given the case, the ground truth, and the candidate's verbatim snippets with source identifiers, decide whether a CS specialist could reconstruct guidance equivalent to the ground truth from those snippets alone. 
	
\section{Skill Code and Judge Prompts}
	\label{app:package}

	This appendix reproduces the skill in full: the operating manual, the tool
	interfaces and output schema, the validator that enforces the evidence contract,
	and the judge prompts used for evaluation. Everything needed to reimplement the
	method on another corpus is here; nothing is deferred to an external artifact. No
	cases, article text, or organization-specific names appear anywhere in it.

	\subsection{\texttt{README.md}}
\begin{lstlisting}
# Skill-based CS-guidance agent — reproducible scaffold (no data)

This is an anonymized scaffold for the skill-based customer-support guidance
agent described in the paper. It contains the **operating manual**, the
**deterministic script interfaces**, the **output validator** (the enforced
"trust boundary"), and the **LLM-judge prompts** — but **no data, no corpus, and
no organization-specific names**. It is sufficient to reproduce the *method* on
any structured knowledge corpus.

## Contents
- `SKILL.md` — the agent operating manual (workflow + constraints + output contract).
- `scripts/INTERFACES.md` — the four deterministic script interfaces the agent calls.
- `scripts/validate_output.py` — the output validator: schema + verbatim-evidence
  enforcement (the trust boundary). Runnable, data-free.
- `judge_prompts.md` — the two evaluation-judge prompts (guidance correctness;
  evidence sufficiency) and the pairwise-preference prompt.

## The idea in one paragraph
A general-purpose LLM coding/agent harness is given (1) a natural-language
operating manual, (2) a few deterministic scripts that load inputs, search a
structured knowledge corpus, fetch source content, and write *validated*
outputs, and (3) the corpus itself organized as a navigation layer over source
articles. The agent composes step-by-step guidance plus verbatim supporting
snippets. A validator script enforces hard constraints (every snippet is a
verbatim substring of its cited source; ground truth is never read) — so
correctness of the output contract does not depend on model goodwill.
\end{lstlisting}

	\subsection{\texttt{SKILL.md}: operating manual}
\begin{lstlisting}
---
name: cs-guidance-agent
description: Produce step-by-step customer-support guidance for a case, with verbatim supporting evidence drawn from a structured knowledge corpus, using a general-purpose agent harness (no bespoke retrieval pipeline). Then evaluate.
---

# CS Guidance Agent (anonymized operating manual)

Produce, for each support case, (a) step-by-step guidance for a human agent and
(b) up to 15 verbatim supporting snippets quoted from source articles. The
knowledge base has two layers:

- **Navigation layer** — a structured, LLM-authored "wiki" of topic pages. Each
  page (i) states *what the underlying workflow solves* (not just its steps),
  (ii) lists the **source articles** it relates to, and (iii) may be a **merged
  entity** synthesizing several articles. Retrieval searches this layer; it is
  what makes "find the right material" easy.
- **Content layer** — the underlying source articles. All verbatim snippets must
  be quoted from here, never from the navigation pages.

## Per-case workflow (identical regardless of retrieval backend)
1. **Load** the case via `load_case.py` (ground truth is stripped — never seek it).
2. **Retrieve** candidates: run 3–5 `retrieve.py "<query>" --case <id>` queries
   covering the primary workflow, sub-topics, and named entities.
3. **Fetch** full bodies of promising candidates via `get_article.py`.
4. **Compose** concise, case-specific step-by-step guidance.
5. **Extract** 10–14 snippets, each copied **verbatim** from a fetched article,
   tagged with its `source_article` id and a one-line `why`.
6. **Save** via `save_output.py` (validates schema + verbatim evidence; rejects
   non-conformant output).

## Hard constraints
- Never read the dataset directly; only `load_case.py`.
- Snippet text must be a verbatim substring of its cited article body.
- ≤ 15 snippets. Do not fabricate citations.
- Do not template guidance; compose per case from that case's retrieved content.

## Configuration
A single config pins corpus paths, output locations, the retrieval backend, and
parameters. Swapping the backend (navigation-layer vs direct-article) changes
only `retrieve.py`; the rest of the workflow is unchanged.
\end{lstlisting}

	\subsection{\texttt{scripts/INTERFACES.md}: tool contract and output schema}
\begin{lstlisting}
# Deterministic script interfaces (the skill's tool contract)

The agent calls exactly four deterministic scripts. They are the skill's trust
boundary and audit surface — the agent reasons, but I/O constraints are enforced
by code.

## `load_case.py --idx N` (or `--id <case_id>`)
Loads one case, **stripping all ground-truth fields** before the agent sees it
(guidance, labels, timestamps). Emits only: `case_id`, `contact_reason`,
`contact_reason_path`, `case_notes`, `case_messages`, `n_messages`, `n_notes`.
→ Guarantees the agent can never read the answer it's evaluated against.

## `retrieve.py "<query>" [-n K] [--case <id>]`
The **swappable retrieval layer**. Returns a ranked list of candidate source
article IDs (+ short preview), regardless of backend. Two reference backends:
- **navigation layer** (recommended): hybrid (BM25+dense) search over a
  structured wiki whose pages cite source-article IDs; return those IDs.
- **direct**: search the source-article corpus directly.
Logs per-call latency. Swapping the backend changes only this script.

## `get_article.py --id <article_id> [--section H] [--summary]`
Reads full source-article content by ID from the corpus (the **content layer**).
The agent quotes verbatim from here. Logs per-call latency.

## `save_output.py --file out.json`
Validates against the output schema and **enforces that every snippet is a
verbatim substring of its cited source article** (see `validate_output.py`);
rejects non-conformant writes. Stamps per-case timing. This is the enforced
contract that makes outputs auditable independent of the model.

## Output schema
```json
{
  "case_id": "<id>", "guidance": "<markdown>",
  "snippets": [{"source_article": "art_NN", "source_title": "...",
                "text": "<verbatim substring of the cited article>", "why": "..."}],
  "n_snippets": <int matching len(snippets)>,
  "articles_consulted": ["art_NN", ...]
}
```
Constraints enforced: ≤15 snippets; every `text` verbatim in its cited article
(modulo unicode-punctuation/whitespace normalization); `source_article` matches
the id scheme; ground truth never loaded.
\end{lstlisting}

	\subsection{\texttt{scripts/validate\_output.py}: the enforced trust boundary}
\begin{lstlisting}[breakatwhitespace=false]
#!/usr/bin/env python3
"""
validate_output.py — the enforced "trust boundary" for a skill-based guidance
agent. Anonymized, data-free. Validates one agent output against the schema and
enforces that every emitted snippet is a verbatim substring of its cited source
article. Rejects non-conformant writes. This is what makes the output contract
auditable independently of the model.

Usage:
    python3 validate_output.py --file out.json --articles articles.json
where articles.json maps {article_id: full_text} for the cited sources.
"""
import argparse, json, re, sys

ARTICLE_RE = re.compile(r"^[a-z]+_\d+$")          # e.g. art_123  (anonymized id scheme)
TOP_REQUIRED = ["case_id", "guidance", "snippets", "n_snippets"]
SNIPPET_REQUIRED = ["source_article", "source_title", "text", "why"]
MAX_SNIPPETS = 15
MIN_SNIPPET_CHARS = 15


def _norm(s: str) -> str:
    """Leniency the validator allows: unicode punctuation + whitespace only."""
    s = (s.replace("\u2019", "'").replace("\u2018", "'")
           .replace("\u201c", '"').replace("\u201d", '"'))
    return re.sub(r"\s+", " ", s).strip()


def validate(output: dict, articles: dict) -> list:
    errors = []
    for f in TOP_REQUIRED:
        if f not in output:
            errors.append(f"missing required field: {f}")
    if errors:
        return errors
    if not isinstance(output["guidance"], str) or not output["guidance"].strip():
        errors.append("guidance must be a non-empty string")
    snippets = output["snippets"]
    if not isinstance(snippets, list) or not snippets:
        errors.append("snippets must be a non-empty array")
        return errors
    if len(snippets) > MAX_SNIPPETS:
        errors.append(f"too many snippets ({len(snippets)} > {MAX_SNIPPETS})")
    for i, s in enumerate(snippets):
        for f in SNIPPET_REQUIRED:
            if not isinstance(s.get(f), str) or not s[f].strip():
                errors.append(f"snippets[{i}].{f} must be a non-empty string")
        aid = s.get("source_article", "")
        if not ARTICLE_RE.match(aid):
            errors.append(f"snippets[{i}].source_article {aid!r} bad id format")
            continue
        text = s.get("text", "")
        if len(text.strip()) < MIN_SNIPPET_CHARS:
            errors.append(f"snippets[{i}].text too short")
            continue
        # THE TRUST BOUNDARY: text must be a verbatim substring of its cited article.
        body = articles.get(aid)
        if body is None:
            errors.append(f"snippets[{i}] cites unknown article {aid}")
        elif _norm(text) not in _norm(body):
            errors.append(f"snippets[{i}] NOT verbatim in cited article {aid} (rejected)")
    if output.get("n_snippets") != len(snippets):
        errors.append("n_snippets does not match len(snippets)")
    return errors


def main():
    ap = argparse.ArgumentParser()
    ap.add_argument("--file", required=True)
    ap.add_argument("--articles", required=True, help="JSON {article_id: full_text}")
    args = ap.parse_args()
    output = json.load(open(args.file))
    articles = json.load(open(args.articles))
    errs = validate(output, articles)
    if errs:
        print("VALIDATION FAILED:", file=sys.stderr)
        for e in errs:
            print("  -", e, file=sys.stderr)
        sys.exit(1)
    print("OK: output conforms to schema and all snippets are verbatim.", file=sys.stderr)


if __name__ == "__main__":
    main()
\end{lstlisting}

	\clearpage
	\subsection{Judge-prompt structure}

	We give each judge as a schematic: the
	slots it is filled with, the order of the reasoning it is asked to perform, and
	the terminal output the scoring code parses. Both are frozen for a run and
	majority-voted over three independent calls.

\begin{lstlisting}
JUDGE 1 - guidance correctness (binary)
 IN   {contact_reason}, {case_notes},
      {candidate_guidance}
      no reference guidance supplied
 ASK  does this guidance resolve the
      case: right procedures, right
      policies, nothing misleading?
 STEP 1) what does the case require?
      2) does the guidance address it?
      3) anything wrong or misleading?
      4) any critical step absent?
 OUT  reasoning, then last line:
      VERDICT <accept|reject>

JUDGE 2 - evidence sufficiency (binary)
 IN   {case}, {reference_guidance},
      {emitted_snippets}
 ASK  could a specialist rebuild
      guidance equivalent to the
      reference from these snippets
      alone?
 CAL  equivalent != identical
      ok: core procedures carried in
          other wording or articles
      not penalised: generic advice,
          minor tips, absent URLs
          where the tool is named
 STEP 1) list reference action points
      2) check each vs the snippets
      3) flag only missing CORE items
 OUT  reasoning, then last line:
      VERDICT <accept|reject>
\end{lstlisting}

	Judge~1 receives no reference, so it scores standalone correctness rather
	than agreement with an approved answer. Judge~2's calibration block exists
	because the judge was otherwise prone to the three failure modes it names. The numbers in this paper come from Judges~1 and~2.

\end{document}